\documentclass[preprint,12pt]{elsarticle}

\usepackage{amssymb}

\usepackage[utf8]{inputenc} 
\usepackage{booktabs}       
\usepackage{amsfonts}       
\usepackage{nicefrac}       
\usepackage{microtype}      
\usepackage{amsmath}
\usepackage{amsthm}
\usepackage{amssymb}
\usepackage{algorithm}
\usepackage{algorithmic}
\usepackage{subfigure}
\usepackage{array}
\usepackage[switch]{lineno}
\usepackage{multirow}
\usepackage{color, xcolor}
\usepackage{url}
\usepackage{ulem}
\usepackage{graphicx}
\usepackage{subfigure}
\usepackage{booktabs} 
\usepackage{multirow}
\usepackage{amsmath}
\usepackage{amssymb}
\usepackage{mathtools}
\usepackage{amsthm}
\usepackage{xcolor,colortbl}
\usepackage{float}
\usepackage{amsthm}

\journal{Neural Networks}

\begin{document}

\begin{frontmatter}



\author[1,2]{Xunyu Zhu}
\ead{zhuxunyu@iie.ac.cn}

\author[1,2]{Jian Li\corref{cor1}}
\ead{lijian9026@iie.ac.cn}

\author[3]{Yong Liu}
\ead{liuyonggsai@ruc.edu.cn}

\author[1,2]{Can Ma}
\ead{macan@iie.ac.cn}

\author[1,2]{Weiping Wang}
\ead{wangweiping@iie.ac.cn}

\cortext[cor1]{Corresponding author}

\affiliation[1]{Institute of Information Engineering, Chinese Academy of Sciences.}

\affiliation[2]{School of Cyber Security, University of Chinese Academy of Sciences.}

\affiliation[3]{Gaoling School of Artificial Intelligence, Renmin University of China.}

\title{Distilling Mathematical Reasoning Capabilities into Small Language Models}

\begin{abstract}
This work addresses the challenge of democratizing advanced Large Language Models (LLMs) by compressing their mathematical reasoning capabilities into sub-billion parameter Small Language Models (SLMs) without compromising performance. We introduce Equation-of-Thought Distillation (EoTD), a novel technique that encapsulates the reasoning process into equation-based representations to construct an EoTD dataset for fine-tuning SLMs. Additionally, we propose the Ensemble Thoughts Distillation (ETD) framework to enhance the reasoning performance of SLMs. This involves creating a reasoning dataset with multiple thought processes, including Chain-of-Thought (CoT), Program-of-Thought (PoT), and Equation-of-Thought (EoT), and using it for fine-tuning. Our experimental performance demonstrates that EoTD significantly boosts the reasoning abilities of SLMs, while ETD enables these models to achieve state-of-the-art reasoning performance.
\end{abstract}




\end{frontmatter}



\section{Introduction}
Large language models (LLMs) like those built on Transformer architectures mark a leap forward in natural language processing. These models, including prominent ones such as LLaMA~\citep{abs-2302-13971}, GPT-4~\citep{abs-2303-08774}, and PaLM~\citep{ChowdheryNDBMRBCSGSSTMRBTSPRDHPBAI23}, boast parameter counts in the hundreds of billions. Trained on vast text datasets, they demonstrate remarkable proficiency in a wide array of downstream tasks.

\begin{figure}[!ht]
	\centering
	\includegraphics[width=10cm]{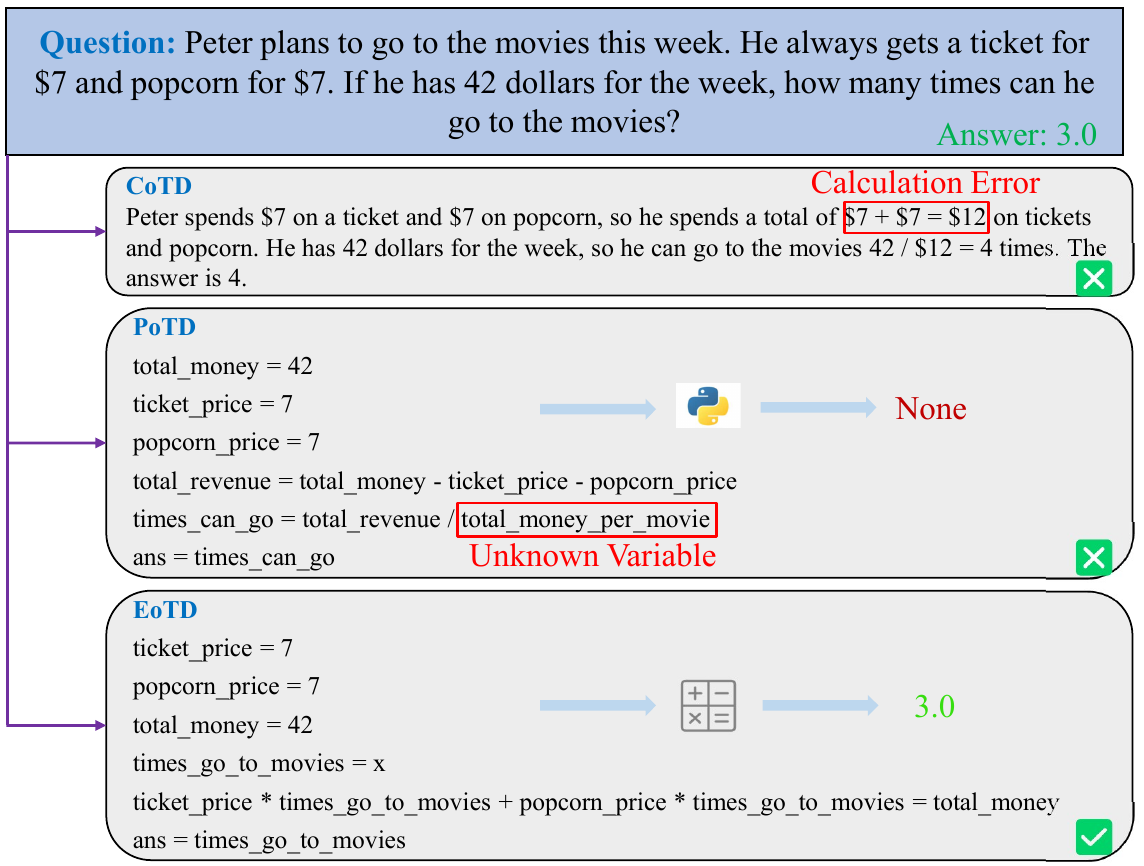}
	\caption{A particular case where SLMs under CoTD and PoTD fail to generate the correct answer, but SLMs under EoTD successfully solve the question.}
	\label{fig:eot_example}
\end{figure}

Recent studies~\citep{chen2023program,0002WSLCNCZ23,wang-etal-2023-plan,liu-etal-2023-plan} have honed the reasoning abilities of LLMs through chain-of-thought (CoT) prompting, which generates intermediate steps to solve complex problems. However, the deployment of such models is challenging due to their size and computational requirements. For example, the GPT-3 model~\citep{NEURIPS2020_1457c0d6} necessitates at least 350GB of FP16 storage and multiple A100 GPUs with 80GB of memory each for efficient inference.

Recent work~\citep{magister-etal-2023-teaching,shridhar-etal-2023-distilling,ho-etal-2023-large,FuPOSK23} investigates distilling LLM reasoning into SLMs (under 1B parameters) for broader deployment. This involves using LLMs to create enriched datasets with detailed reasoning paths, which then fine-tune SLMs, endowing them with advanced reasoning abilities.  For example, Chain-of-Thought Distillation (CoTD)~\citep{ho-etal-2023-large} encapsulates the reasoning process into textual rationales, and Program-of-Thought Distillation (PoTD)~\cite{abs-2305-13888} formulates the reasoning process into python program. The methodologies for CoTD and PoTD are delineated in~\ref{sec:CoTD} and~\ref{sec:PoTD}, respectively. These distillation methods have distinct strengths and limitations. As illustrated in Figure~\ref{fig:eot_example}, CoTD instructs SLMs to generate a step-by-step reasoning flow in natural language and perform calculations in real-time, offering a flexible solution format but risking incorrect answers due to calculation errors~\cite{ho-etal-2023-large}. PoTD addresses this issue by training SLMs to conceptualize questions as programs, which are then executed by a Python interpreter to produce the final answer. However, SLMs under PoTD sometimes generate unknown variables, but the python interpreter can execute programs only if every variable of programs is defined with a value.  Another intriguing method involves framing math problems as linear equation systems~\cite{liu-etal-2023-plan}. Inspired by this, we introduce Equation-of-Thought Distillation (EoTD), teaching SLMs to model math questions in a similar manner. EoTD facilitates a direct understanding of mathematical principles and fosters logical thinking in SLMs.

The diversity within each distillation method doesn't signify rivalry or exclusivity. Instead, in practical problem-solving situations, employing multiple methods can offer complementary advantages. These distinct distillation approaches can synergize, resulting in benefits that exceed those achievable with any single approach. Motivated by this, we introduce Ensemble Thoughts Distillation (ETD) to further enhance SLM reasoning. ETD merges the CoTD, PoTD, and EoTD datasets into a comprehensive ETD dataset for fine-tuning SLMs. The diversity of reasoning strategies within the ETD dataset enriches the reasoning knowledge, contributing to its effectiveness.

We assessed EoTD and ETD across CodeT5 models from Small (0.06B) to Large (0.77B) on four mathematical reasoning datasets. Results indicate EoTD significantly boosts SLM reasoning abilities, while ETD enables SLMs to reach state-of-the-art (SOTA) performance. For instance, with EoTD, CodeT5-Small reached 18.87\% accuracy on GSM8K, and ETD elevated CodeT5-Large to 42.45\% accuracy. Ablation studies confirm that the volume and variety of reasoning paths in ETD correlate with improved SLM reasoning performance.

\section{Related Work}
\subsection{Large Language Models (LLMs)} 
Building on the insights from~\cite{wei2022chain, huang-chang-2023-towards, wei2022finetuned}, our research investigates the distillation of complex reasoning pathways from LLMs, such as GPT-4~\citep{abs-2303-08774} and PaLM-2~\citep{abs-2305-10403}, into more manageable models. These LLMs, with their vast parameter counts exceeding 100 billion, have demonstrated a profound capacity for navigating intricate reasoning tasks. They can independently construct a series of logical steps leading to a conclusion, particularly when provided with structured reasoning examples or when guided by prompts that encourage stepwise thinking. Our work aims to capture this advanced reasoning in smaller models, thus reducing the computational overhead and making such capabilities more widely accessible.

The formidable reasoning skills of LLMs on complex tasks are offset by their extensive size and computational demands. Deploying models like GPT-3 \citep{abs-2005-14165} for inference, for example, demands at least 320GB of storage for FP16 parameters and no fewer than five A100 GPUs with 80GB of memory each for efficient functioning. These requirements pose significant challenges, particularly for resource-limited settings. Our research addresses these issues by distilling the reasoning capabilities of LLMs into smaller, more computationally feasible models.


Our work addresses these limitations by focusing on the distillation of reasoning abilities from LLMs into smaller models. This process aims to retain the advanced reasoning capabilities of LLMs while significantly reducing resource requirements. Consequently, our approach facilitates the democratization of cutting-edge NLP technologies, enabling the use of powerful reasoning tools in settings with constrained computational resources.

\subsection{Mathematical Reasoning}
Mathematical Reasoning tasks, highlighted by benchmarks like GSM8K~\citep{abs-2110-14168} and SVAMP~\citep{patel-etal-2021-nlp}, pose a significant challenge for LLMs. To improve LLMs' performance in this domain, researchers have pinpointed two main strategies.


\textbf{Chain-of-Thought Reasoning}\quad
LLMs' reasoning can be enhanced by prompting them to articulate intermediate steps towards a solution, as demonstrated by Wei et al.~\cite{wei2022chain}. This insight has led to various advancements~\citep{chen2023program,0002WSLCNCZ23,wang-etal-2023-plan,liu-etal-2023-plan} that refine reasoning paths. For example, Chen et al.~\cite{chen2023program} prompt LLMs to generate executable code, Wang et al.~\cite{0002WSLCNCZ23} use multiple reasoning paths with a voting mechanism for the correct answer, Wang et al.~\cite{wang-etal-2023-plan} have LLMs create a plan before reasoning, and Liu et al.~\cite{liu-etal-2023-plan} employ diverse reasoning prompts for problem-solving. Building on these methods, our work introduces Equation-of-Thought Distillation (EoTD) to further improve SLMs' mathematical reasoning.

\textbf{Finetuning-based Reasoning}
refines LLMs like Llama2~\citep{abs-2307-09288}, Qwen~\citep{abs-2309-16609}, and Baichuan2~\citep{abs-2309-10305} by drawing on techniques from advanced models such as GPT-4~\citep{abs-2303-08774} and PaLM-2~\citep{abs-2305-10403}. Notably, Yuan et al.~\cite{abs-2308-01825} employ Rejection Sampling Fine-Tuning (RFT) to enhance LLMs' mathematical reasoning, while WizardMath\citep{abs-2308-09583} uses Reinforcement Learning from Evolved Instructions Feedback (RLEIF) to improve LLaMA-2's reasoning abilities. MAmmoTH~\citep{abs-2309-05653} combines CoT and PoT rationales for more effective instruction-tuning of LLMs in math problem-solving. Despite their effectiveness, the large size of these LLMs limits their deployment efficiency.

\subsection{Knowledge Distillation}
Knowledge Distillation optimizes LLMs for practical use by transferring knowledge from larger models to smaller, efficient ones~\citep{abs-2308-07633}. Research~\citep{magister-etal-2023-teaching,shridhar-etal-2023-distilling,ho-etal-2023-large,FuPOSK23} has aimed to endow compact models like T5~\citep{RaffelSRLNMZLL20} and GPT-2~\citep{radford2019language} with the advanced reasoning of LLMs such as GPT-4~\citep{abs-2303-08774} and PaLM-2~\citep{abs-2305-10403}. For example, Ho et al.~\cite{ho-etal-2023-large} fine-tune student models using the most accurate reasoning paths from LLMs. Shridhar et al.~\cite{shridhar-etal-2023-distilling} train a dual-model system on sub-questions and solutions, while Fu et al.~\cite{FuPOSK23} suggest scaling down general competencies of smaller models to boost task-specific performance. Our work presents a novel distillation approach that encodes mathematical reasoning as equations and introduces Ensemble Thoughts Distillation, combining CoT, EoT, and PoT to create a diverse dataset with more abundant reasoning knowledge. Our results demonstrate state-of-the-art performance in mathematical reasoning.

\begin{figure*}[!ht]
	\centering
	\includegraphics[width=\textwidth]{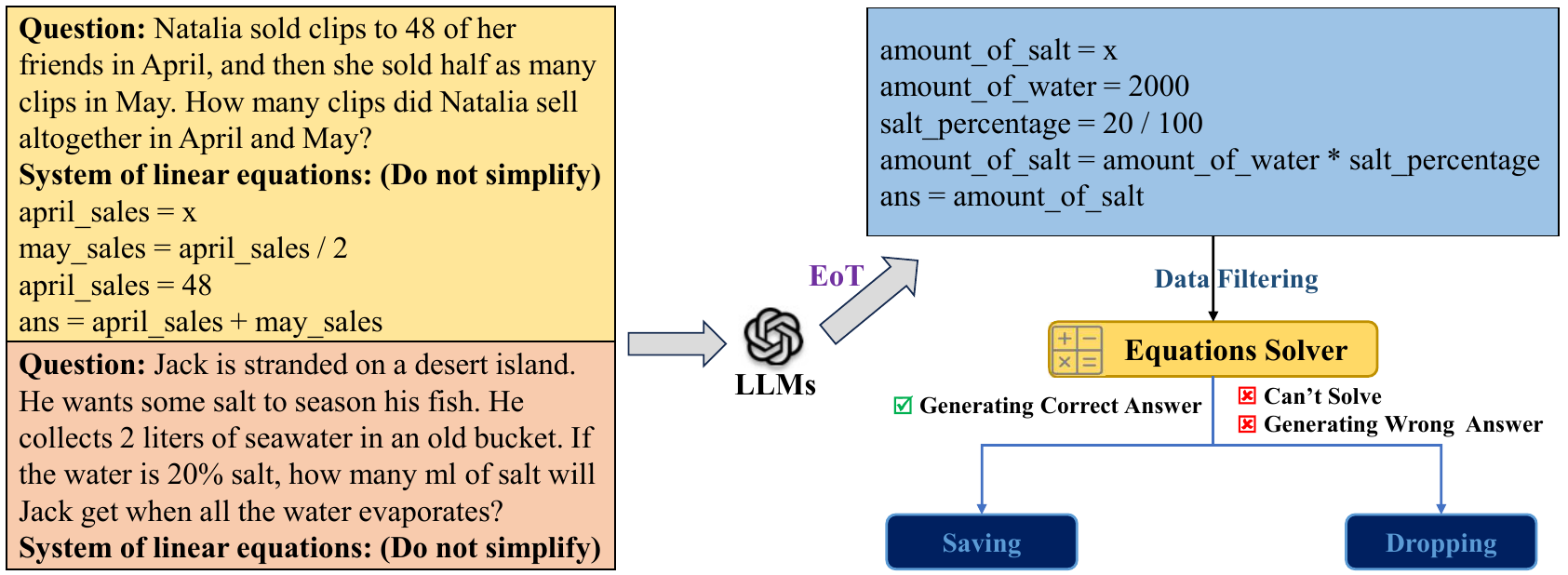}
	\caption{\textbf{Detailed data generation of our framework.} Firstly, we manually construct some contextualized examples, and combine these contextualized examples, the question, and the prompt ``System of linear equations: (Do not simplify)" to prompt LLMs to generate EoT based on the question. This equations system is sent to  a deterministic equation solver, if there are compile errors or if it produces wrong answer, we will drop the EoT. Finally, we get a high-quality reasoning dataset.}
	\label{fig:eot_data_generation}
\end{figure*}

\section{Methodology}
In this work, we introduce a novel distillation method for mathematical reasoning tasks, termed Equation-of-Thought Distillation (EoTD), which translates mathematical reasoning into equations for fine-tuning SLMs.

\subsection{Equation-of-Thought Distillation}
\label{subsec:eotd}
\subsubsection{Data Generation from LLMs}
\label{subsec:eotdg}
Our EoTD framework commences by creating a dataset from LLMs, which precedes SLM fine-tuning. As illustrated in Figure~\ref{fig:eot_data_generation}, we employ in-context learning \citep{abs-2301-00234,min-etal-2022-rethinking,rubin-etal-2022-learning} to prompt LLMs for reasoning data. Within a mathematical dataset $\mathcal{D}$, each entry $(x,y)$ pairs a question $x$ with its answer $y$. We select $k$ samples $\{(x_1, y_1), \ldots, (x_k, y_k)\}$ from $\mathcal{D}$ and manually craft rationales $e$ in EoT format. These form contextualized instances $\{(x_1, e_1, y_1), \ldots, (x_k, e_k, y_k)\}$, compiled into a demonstration set $\mathcal{D}_D$. We then prompt LLMs with a question and the instruction ``System of linear equations: (Do not simplify)" to generate rationales. The EoT generation is formalized as:
\begin{equation*}
	e_i = f_\mathcal{M}(x_i, \mathcal{D}_D),
\end{equation*}
where $\mathcal{M}$ denotes the LLM, $f$ the decoding function, and $i$ the index in $\mathcal{D}$. This yields the EoT dataset $\mathcal{D}_E$, composed of triplets $(x, e, y)$.

\textbf{Data Filtering}---After LLMs generate EoT dataset, we validate each equation system using an external Equation Solver to ensure the accuracy of our initial dataset $\mathcal{D}_E$. Any equation system that cannot be solved or produces an incorrect result is excluded. This rigorous filtering process removes errors, enhancing the dataset's quality. Consequently, this refinement directly improves the performance of fine-tuned SLMs by providing cleaner and more reliable training data.

\begin{figure*}[!ht]
	\centering
	\includegraphics[width=\textwidth]{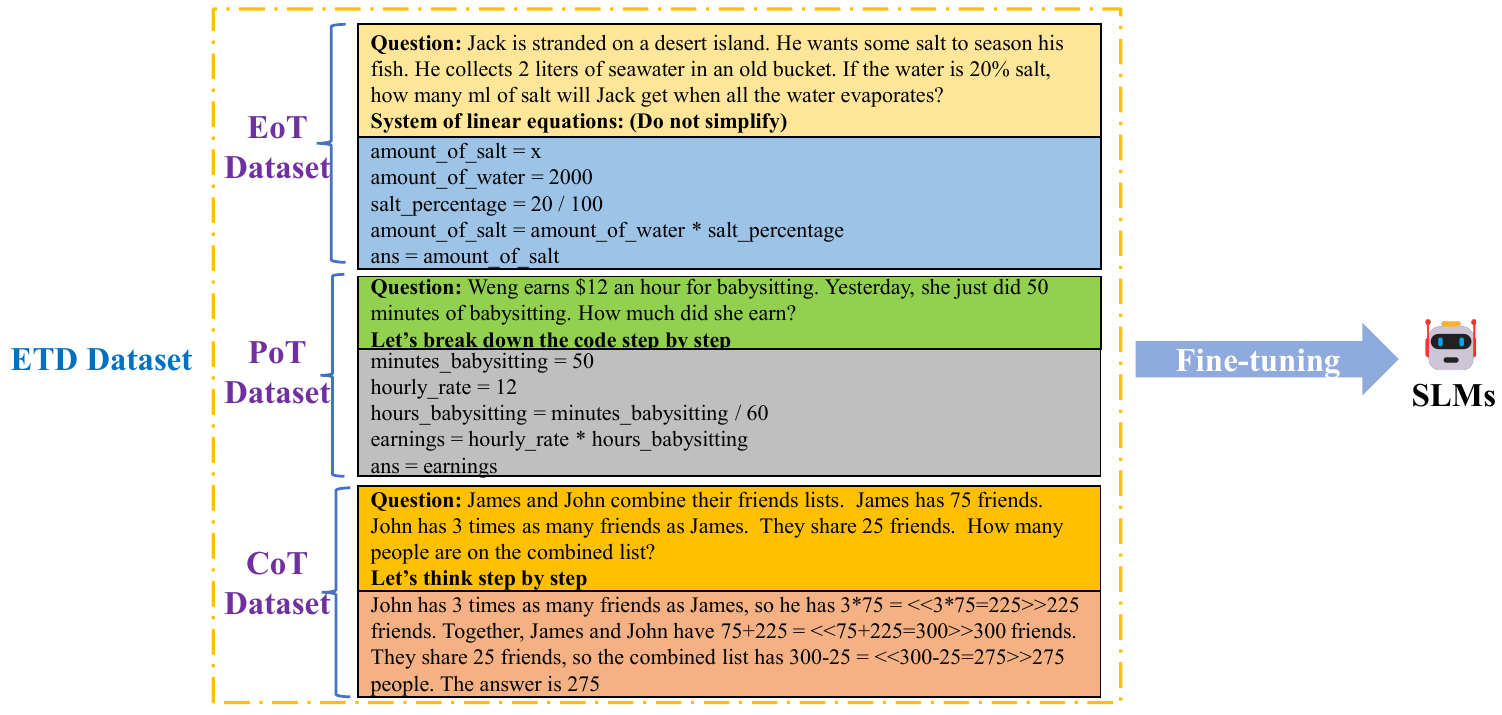}
	\caption{\textbf{Detailed overview of Ensemble Thought Distillation.} Firstly, we combine a CoT dataset, a PoT dataset and a EoT dataset to build a new ETD dataset. The ETD dataset has diverse thoughts and prompts. Then, we use the ETD dataset to fine-tune SLMs. After fine-tuning, we use the prompt ``System of linear equations: (Do not simplify)" to instruct SLMs to generate equations, the prompt ``Let’s break down the code step by step" to instruct SLMs to generate programs, and the prompt ``Let's think step by step" to instruct SLMs to generate chains to solve questions. }
	\label{fig:ETD}
\end{figure*}

\subsubsection{Fine-tuning SLMs}

After assembling the reasoning dataset $\mathcal{D}_E$, we fine-tune SLMs on it. For each training instance $(x, e, y)$ from $\mathcal{D}_E$, we prepend the prompt $p_e$ ``System of linear equations: (Do not simplify)" to the question $x$ and employ fine-tuning to guide the SLM in generating the corresponding equations. The fine-tuning loss function is:
\begin{equation*}
	\mathcal{L} = - \sum_{i=1}^{N} \sum_{t=1}^{T} \log P(e^i_t \mid e^i_{< t}, x^i, p_e),
\end{equation*}
where $N$ is the number of examples in $\mathcal{D}_E$, $p_e$ is the prompt, and $e_{:T}$ is the sequence of equations. Post fine-tuning, the SLM can generate equations for complex questions, which are then solved by an external Equation Solver to obtain the final answer.

\subsection{Ensemble Thoughts Distillation}

ETD merges the CoTD, PoTD, and EoTD to improve diversity of reasoning forms. The diverse reasoning forms have a richer knowledge of reasoning, and can be effective to improve reasoning ability of SLMs. We now detail the ETD method and its implementation.

In parallel with EoTD, we construct a Program-of-Thought (PoT) dataset $\mathcal{D}_p$, with each entry formatted as a triplet $(x, p, y)$. The PoT data generation mirrors the EoT process described in Section~\ref{subsec:eotdg}. We utilize in-context learning to prompt LLMs to generate programmatic solutions for given questions. These programs are then executed by an external Python interpreter to obtain the final answers. Programs that fail to compile or yield incorrect answers are discarded, ensuring the PoT dataset $\mathcal{D}_p$ is of high quality. Similarly, we compile a Chain-of-Thought (CoT) dataset $\mathcal{D}_c$, with each instance also structured as a triplet $(x, c, y)$. The methodologies for CoTD and PoTD are delineated in~\ref{sec:CoTD} and~\ref{sec:PoTD}, respectively.

As depicted in Figure~\ref{fig:ETD}, we amalgamate the EoT, CoT, and PoT datasets to form the new ETD dataset $\mathcal{D}_{ETD}$. For EoT entries, we append the prompt $p_e$ ``System of linear equations: (Do not simplify)" to each question. For PoT entries, we add the prompt $p_r$ ``Let’s break down the code step by step," and for CoT entries, we include the prompt $p_c$ ``Let’s think step by step." These prompts are designed to guide the generation of thoughts in their respective formats. The combined datasets, now enriched with diverse thoughts and instructions, constitute the ETD dataset. We then apply fine-tuning to fine-tune SLMs on $\mathcal{D}_{ETD}$. The loss function for this fine-tuning process is:
\begin{equation*}
	\mathcal{L} = - \sum_{i=1}^{N} \sum_{t=1}^{T} \log P(r^i_t \mid r^i_{< t}, x^i),
\end{equation*}
where $x$ is the input comprising both the question and its associated prompt, and $r$ represents the generated thoughts conditioned on the input. This approach aims to enhance the SLMs' ability to process and generate a variety of thought patterns, thereby improving their mathematical reasoning performance.

After fine-tuning, the SLMs are primed to generate different types of reasoning outputs based on the given prompts. When presented with a question, the prompt ``System of linear equations: (Do not simplify)" elicits equation generation, ``Let’s break down the code step by step" induces program generation, and ``Let’s think step by step" prompts the creation of chains of thought. The SLMs' outputs are then used to derive the final answers. As evidenced in Section~\ref{sec:ETD_effect}, PoT outperforms the others in reasoning, with EoT in second place, and CoT trailing. Consequently, the PoT-generated answer is initially considered as the final answer. If the PoT-generated program fails to compile correctly, the SLMs' EoT-generated answer is then evaluated. Should the EoT-generated equations be unsolvable, the CoT-generated result is finally considered. This hierarchical approach to reasoning ensures the most reliable answer is selected. The detailed reasoning process of ETD is illustrated in Figure~\ref{inference_ETD}.

\section{Experiments}
\label{sec:exp}
\subsection{Dataset}
Our training dataset is derived from the GSM8K~\citep{abs-2110-14168} training set. We construct separate EoTD, PoTD, and CoTD datasets, which are then amalgamated to form the ETD dataset. This ensemble dataset features a variety of prompts and thought processes, on which we fine-tune SLMs. The mathematical reasoning capabilities of the SLMs are evaluated using the GSM8K~\citep{abs-2110-14168} test set, as well as ASDiv~\citep{miao-etal-2020-diverse}, SVAMP~\citep{patel-etal-2021-nlp}, and MultiArith~\citep{roy-roth-2015-solving}.

\begin{table}[]
	\centering
	\scriptsize
	\begin{tabular}{@{}cccc@{}}
		\toprule
		\textbf{Method} & \textbf{Sampled Dataset} & \textbf{Filtered Dataset} & \textbf{Drop Rate(\%)} \\ \midrule
		CoTD & 29892 & 24392 & 18.4 \\
		PoTD & 29892 & 22491 & 24.8 \\
		EoTD & 29892 & 15946 & 46.7 \\ \bottomrule
	\end{tabular}
	\caption{\textbf{Size of the datasets used in our experiments.} Drop Rate refers to the size of dropped data divided by the size of the sampled data.}
	\label{tab:data-size}
\end{table}

\subsection{Implementation}
We employ ChatGPT (gpt-3.5-turbo) as the teacher LLM to construct our training dataset and utilize CodeT5 models—Small (60M), Base (220M), and Large (770M)~\citep{wang-etal-2021-codet5}—as student SLMs. We manually create 8 examples to guide ChatGPT in generating 4 reasoning paths for each dataset (EoT, PoT, and CoT), and Table~\ref{tab:data-size} shows the size of the datasets used in our experiments. Fine-tuning of all student SLMs is conducted using the Huggingface library~\cite{wolf-etal-2020-transformers} on an NVIDIA 3090 GPU with 24 GB RAM. The learning rate for fine-tuning is set to 5e-4, with a total of 10 fine-tuning epochs.

\subsection{Baselines}
\textbf{Proprietary Large Language Models}\quad We present CoT prompting results from an array of SoTA LLMs, such as OpenAI's GPT-4, ChatGPT (gpt-3.5-turbo), Google’s PaLM-2, and Anthropic’s Claude-2.

\textbf{Open-Source Large Language Models}\quad We present  mathematical reasoning performance of Llama-2-7B, CodeLLaMA-7B, and their fine-tuned versions, such as Platpus-2, WizardMath, TORA.

\textbf{Fine-tuned Small Language Models}\quad We present some works that try to fine-tune SLMs under 1B, such as Ho et al.~\cite{ho-etal-2023-large} fine-tune GPT-3-ada, Fu et al.~\cite{FuPOSK23} fine-tune FlanT5, and Shridhar et al.~\cite{shridhar-etal-2023-distilling} fine-tune GPT-2.

\begin{table*}[!ht]
	\centering
	\scriptsize
	\begin{tabular}{c|c|cccc|c}
		\toprule 
		\textbf{Models} & \textbf{\#Params} & \textbf{GSM8K} & \textbf{ASDiv} & \textbf{SVAMP} & \textbf{MultiArith} & \textbf{AVG}\\
		\midrule 
		\rowcolor[rgb]{0.93,0.93,0.93}
		\multicolumn{7}{l}{\textit{Proprietary Large Language Models}} \\
		GPT-4~\citep{abs-2303-08774} & - & 92.0 & 91.3 & 93.1 & - & 92.13\\
		ChatGPT & -  & 80.8 & 87.3 & 83.0 & - & 83.7\\
		Claude-2~\citep{Claude-2} & - & 85.2 & - & - & - & 85.2\\
		PaLM-2~\citep{abs-2305-10403} & 540B & 80.7 & - & - & - & 80.7\\
		\midrule 
		\rowcolor[rgb]{0.93,0.93,0.93}
		\multicolumn{7}{l}{\textit{Open-Source Large Language Models}} \\
		Llama-2~\citep{abs-2307-09288} & 7B & 13.3 & 50.7 & 38.0 & - & 34\\
		CodeLLaMA~\citep{abs-2308-12950} & 7B & 34.0 & 61.4 & 59.0 & - & 51.46\\
		Platypus-2~\citep{abs-2308-07317} & 7B & 14.4 & 47.9 & 36.7 & - & 33\\
		WizardMath~\citep{abs-2308-09583} & 7B & 54.9 & 59.1 & 57.3 & - & 57.1\\
		TORA~\citep{abs-2309-17452} & 7B & 68.8 & 73.9 & 68.2 & - & 70.3 \\
		\midrule 
		\rowcolor[rgb]{0.93,0.93,0.93}
		\multicolumn{7}{l}{\textit{Fine-tuned Small Language Models}} \\
		Ho et al.~\cite{ho-etal-2023-large} & 0.3B & 3.11 & - & - & - & 3.11\\
		Fu et al.~\cite{FuPOSK23} & 0.76B & 20.2 & 23.8 & 20.4 & 38.5 & 25.72\\
		Fu et al.~\cite{FuPOSK23} & 0.25B & 13.4 & 20.9 & 14.2 & 29.7 & 19.55\\
		Shridhar et al.~\cite{shridhar-etal-2023-distilling} & 0.77B & 17.89 & - & 18.14 & - & 18.01\\
		Zhu et al.~\cite{abs-2305-13888} & 0.77B & 39.2 & 51.2 & 48.2 & 79.2 & 54.45\\
		\midrule 
		\rowcolor[rgb]{0.93,0.93,0.93}
		\multicolumn{7}{l}{\textit{Our fine-tuned Small Language Models}} \\
		CodeT5-Small & 0.06B & 1.1 & 0.3 & 0.2 & 0.6 & 0.55\\
		(+) EoTD & & 18.87 & 29.24 & 31.5 & 24.66 & 26.06\\
		(+) ETD & & \textbf{33.58} & \textbf{49.09} & \textbf{42.8} & \textbf{67.83} & \textbf{48.14}\\
		\hline
		CodeT5-Base & 0.22B & 0.8 & 0.2 & 0.0 & 0.0 & 0.25\\
		(+) EoTD & & 27.21& 38.26 & 38.8 & 41.66 & 36.48\\
		(+) ETD & & \textbf{40.63} & \textbf{51.66} & \textbf{48.8} & \textbf{81} & \textbf{55.52}\\
		\hline
		CodeT5-Large & 0.77B & 2.9 & 3.6 & 0.0 & 0.0 & 1.62\\
		(+) EoTD & & 33.13 & 44.03 & 46.1 & 57.33 & 45.14\\
		(+) ETD & & \textbf{42.45} & \textbf{52.81} & \textbf{49.59} & \textbf{85.5} & \textbf{57.58}\\
		\bottomrule 
	\end{tabular}
	\caption{\textbf{Overall test set performance.} We use EoTD and ETD to fine-tune SLMs, and evaluate them on four mathematical reasoning datasets, i.e., GSM8K, ASDiv, SVAMP, and MultiArith. The experiment results show that EoTD can effectively improve SLMs' reasoning performance, and ETD makes SLMs achieve SOTA reasoning performance. }
	\label{tab:main_results}
\end{table*}

\subsection{Main Results}

Table~\ref{tab:main_results} showcases our method's performance on four mathematical datasets, revealing key insights: (1) EoTD significantly enhances the mathematical reasoning of SLMs, with absolute improvements ranging from 25.51\% to 43.52\% across tasks. The key difference between EoTD and baseline approaches lies in the form of reasoning. Baselines typically rely on CoTD, which involves generating numerous steps and performing extensive calculations. While CoT has been shown to significantly improve the mathematical reasoning of LLMs, it is challenging for SLMs due to their limited capacity. In contrast, EoTD delegates the computational tasks to an Equation Solver, allowing the model to focus solely on generating reasoning steps. This approach significantly reduces the cognitive load on the model. Moreover, EoTD's generation of equations provides a more intuitive representation of the relationships between variables, aiding SLMs in understanding and analyzing problems. Therefore, compared to CoTD, EoTD is better suited to improving the mathematical reasoning ability of SLMs.
	(2) ETD outperforms previous state-of-the-art fine-tuned SLMs at all scales, with absolute improvements between 48.14\% and 57.58\% across tasks. Furthermore, ETD's accuracy is approximately 20\% higher than that of EoTD, underscoring the advantage of diverse prompts and thoughts in bolstering SLMs' reasoning capabilities. Earlier distillation datasets predominantly used a single form of reasoning, limiting SLMs to learning fragmented reasoning knowledge. However, different reasoning forms have distinct focal points: CoT offers clear intermediate steps, making reasoning processes easier to understand and interpret, thus increasing transparency and credibility. EoT, based on mathematical principles and formulas, ensures high rigor and accuracy. PoT allows for reasoning automation via programming, boosting reasoning efficiency and accuracy. By generating datasets with multiple reasoning forms, SLMs can learn mathematical reasoning from various perspectives, thereby improving their overall mathematical reasoning capabilities.
	(3) Model size is crucial for reasoning distillation efficacy in SLMs; larger models assimilate more reasoning knowledge, translating to superior performance. For instance, under ETD, CodeT5-Small attains 33.58\% accuracy on GSM8K, CodeT5-Base reaches 40.63\%, and CodeT5-Large achieves 42.45\%.

\subsection{ETD Enhances Thoughts Distillation}
\label{sec:ETD_effect}
\begin{figure}[!ht]
	\centering
	\includegraphics[width=\columnwidth]{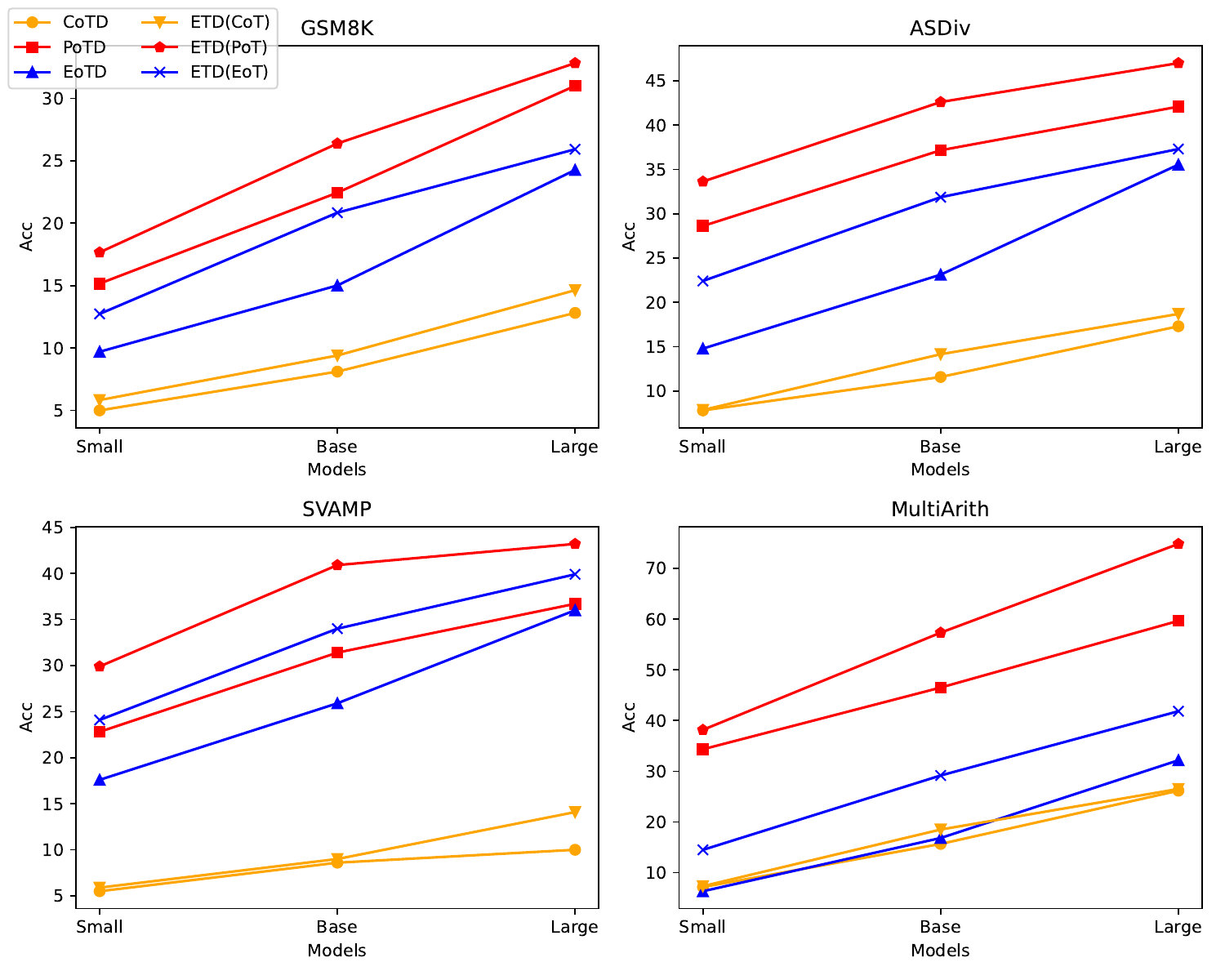}
	\caption{\textbf{Effect of ETD.} We fine-tune SLMs on the ETD dataset, the CoTD dataset, the PoTD dataset, and the EoTD dataset to study the effect of ETD. The experiment results shows that ETD  can improve reasoning performance of SLMs under different thoughts.}
	\label{fig:ETD_effect}
\end{figure}

In this subsection, we examine if ETD can enhance the reasoning abilities of SLMs across different thought processes. Initially, we generate CoTD, PoTD, and EoTD datasets from the GSM8K training set, each containing one reasoning path per question. These datasets are then merged to create the ETD dataset. Subsequently, we fine-tune SLMs, including CodeT5-Small, Base, and Large, using the ETD dataset. The reasoning performance of these SLMs is assessed on the GSM8K test dataset, as well as ASDiv, SVAMP, and MultiArith.

Figure~\ref{fig:ETD_effect} illustrates the outcomes of our experiments, from which we deduce that: (1) ETD enhances SLMs' reasoning performance. SLMs fine-tuned with ETD outperform those trained on CoTD, PoTD, and EoTD in CoT, PoT, and EoT tasks, respectively. For instance, CodeT5-Base achieves a 26.38\% PoT accuracy on the GSM8K test dataset under ETD, surpassing the 22.44\% PoT accuracy under PoTD. Similarly, CodeT5-Small reaches a 24.09\% EoT accuracy on SVAMP with ETD, compared to 17.59\% with EoTD. (2) SLMs gain more valuable reasoning knowledge from PoTD, reflected in superior reasoning performance, with EoTD and CoTD following in that order. This pattern persists with ETD, leading to a hierarchical approach where the PoT-generated answer is preferred, followed by EoT if PoT fails, and CoT as a last resort. (3) Scaling up the size of student models consistently improves the performance across CoTD, PoTD, EoTD, and ETD, indicating that larger models benefit more from our method.

\subsection{The Effect of Different Thoughts in ETD}
\begin{table}[!ht]
	\centering
	\scriptsize
	\begin{tabular}{c|cccc}
		\toprule 
		\textbf{Methods}  & GSM8K & ASDiv & SVAMP & MultiArith\\
		\midrule
		CoTD & 8.11 & 11.59 & 8.6 & 15.66 \\
		PoTD & 22.44 & 37.16 & 31.4 & 46.5 \\
		EoTD & 15.01 & 23.13 & 25.90 & 16.83 \\
		\midrule
		\rowcolor[rgb]{0.93,0.93,0.93}
		\multicolumn{5}{l}{\textit{CoTD + PoTD}} \\
		CoT & 7.96 & 13.12 & 11.3 & 14.16 \\
		PoT & 25.01 & 39.59 & 35.6 & 53.16 \\
		\midrule
		\rowcolor[rgb]{0.93,0.93,0.93}
		\multicolumn{5}{l}{\textit{CoTD + EoTD}} \\
		CoT & 8.49 & 13.02 & 8.79 & 15.16\\
		EoT & 17.13 & 26.62 & 28.19 & 20.5\\
		\midrule
		\rowcolor[rgb]{0.93,0.93,0.93}
		\multicolumn{5}{l}{\textit{PoTD + EoTD}} \\
		PoT & 26.23 & 39.59 & 34.8 & 59 \\
		EoT & 18.65 & 29.10 & 31.5 & 25.5\\
		\midrule
		\rowcolor[rgb]{0.93,0.93,0.93}
		\multicolumn{5}{l}{\textit{CoTD + PoTD + EoTD}} \\
		CoT & 9.4 & 14.16 & 9 & 18.5 \\
		PoT & 26.38 & 42.6 & 40.9 & 57.33 \\
		EoT & 20.84 & 31.87 & 34 & 29.16\\
		\bottomrule
	\end{tabular}
	\caption{\textbf{Effect of Thoughts in ETD.} We fine-tune CodeT5-Base on ETD datasets which own different thoughts in it to analyse the effect of thoughts in ETD. The experiment results show that when the ETD dataset own more thoughts, SLMs fine-tuned on it can achieve better reasoning performance.}
	\label{tab:effect_different_thoughts}
\end{table}


In this subsection, we investigate the impact of different reasoning paths within the ETD framework. We begin by fine-tuning CodeT5-Base on individual datasets—CoTD dataset, PoTD dataset, and EoTD dataset—each containing a unique reasoning path per question. We then extend our fine-tuning to combinations of these datasets: CoTD with PoTD dataset, CoTD with EoTD dataset, PoTD with EoTD dataset, and the full ETD dataset which integrates CoTD, PoTD, and EoTD. This approach allows us to assess the influence of each reasoning path and their synergistic effects on the model's performance.


Table~\ref{tab:effect_different_thoughts} presents the results of our experiments, from which we observe that: (1) SLMs exhibit improved reasoning performance with an increasing number of thought processes incorporated into ETD. For instance, CodeT5-Base, when trained on CoTD and PoTD combined, achieves a 25.01\% PoT accuracy on the GSM8K test dataset, which further increases to 26.38\% when trained on the full combination of CoTD, PoTD, and EoTD. (2) SLMs trained on the PoTD and EoTD combination outperform those trained on either the CoTD and PoTD or the CoTD and EoTD combinations. This suggests that the structured nature of PoTD and EoTD contributes to SLMs' ability to assimilate more valuable knowledge effectively.

\subsection{More Data Improves Reasoning Ability of SLMs}
\begin{figure}[!ht]
	\centering
	\includegraphics[width=\columnwidth]{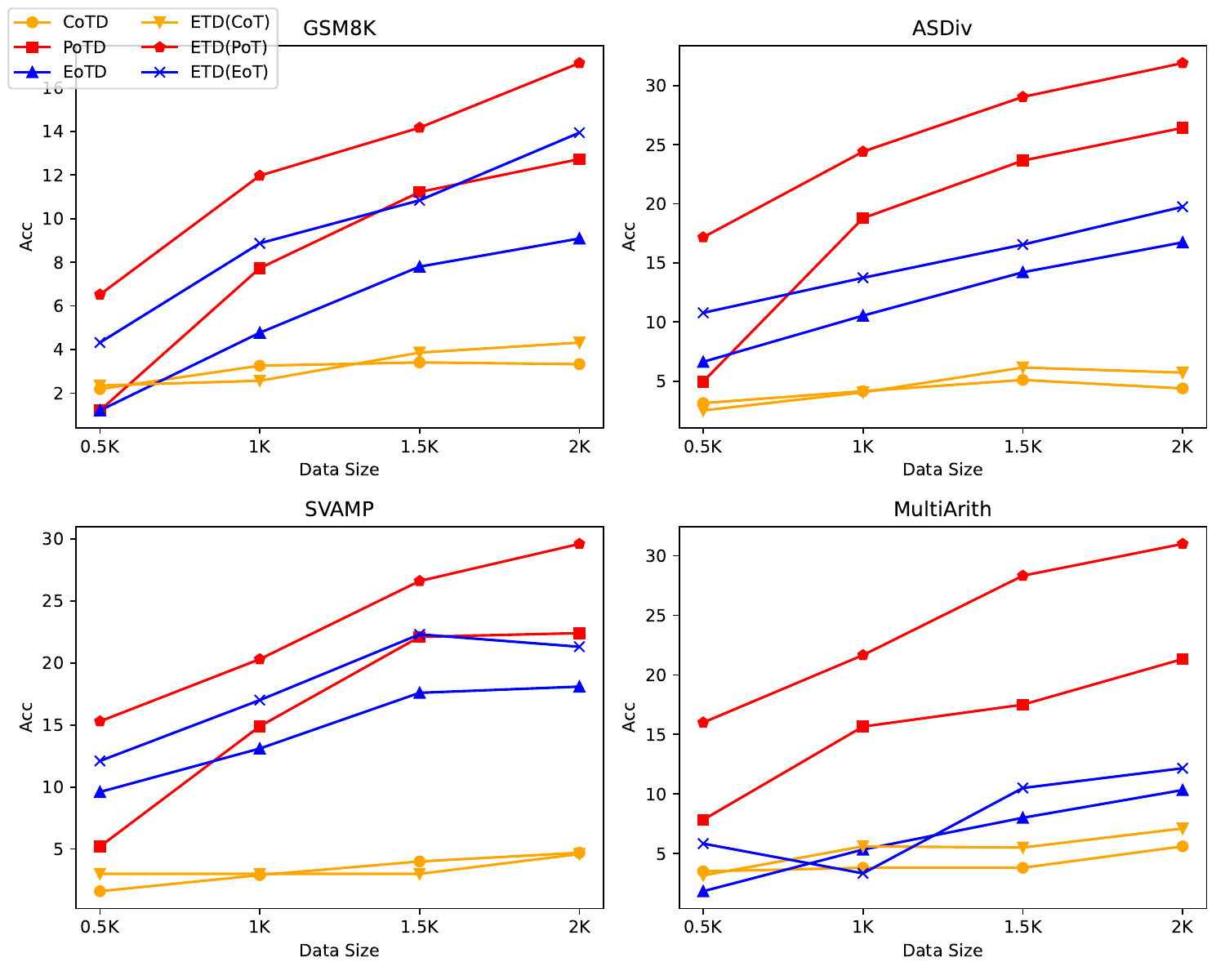}
	\caption{\textbf{Effect of Data Scale.} We fine-tune CodeT5-Base under different data sizes to evaluate the effect of data scale. The experiment results show that larger data size make SLMs better reasoning performance.}
	\label{fig:data_size_effect}
\end{figure}


In this subsection, we explore the influence of dataset size on the reasoning capabilities of SLMs. We create subsets of varying sizes from our reasoning datasets and utilize these to fine-tune CodeT5-Base. This analysis helps determine the relationship between the amount of data and the model's performance in reasoning tasks.


Figure~\ref{fig:data_size_effect} depicts the outcomes of our experiments, indicating that larger datasets enhance the reasoning performance of SLMs. For instance, CodeT5-Base, when fine-tuned on a 1K ETD dataset, attains a 24.42\% PoT accuracy on the ASDiv test set, whereas training on a smaller 0.5K ETD dataset results in a lower PoT accuracy of 17.17\% on the same test set. This trend underscores the positive correlation between dataset size and the model's reasoning proficiency.

\subsection{Diverse Reasoning Paths Improve SLMs' Reasoning Performance}
\begin{figure}[!ht]
	\centering
	\includegraphics[width=\columnwidth]{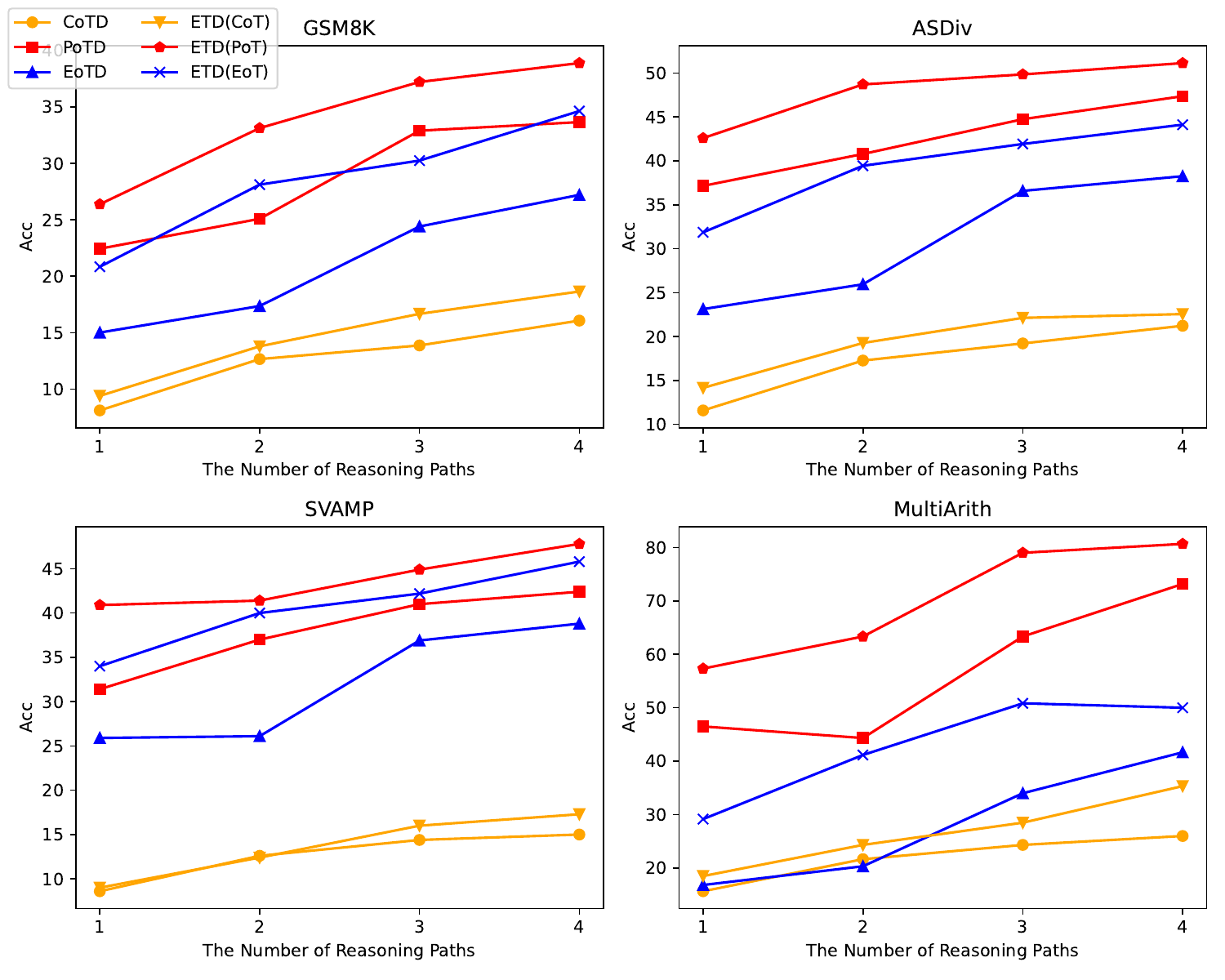}
	\caption{\textbf{Effect of Reasoning Paths.} We fine-tune CodeT5-Base with different reasoning paths to analyse the effect of reasoning paths. The experiment results shows that diverse reasoning paths can improve SLMs' reasoning performance.}
	\label{fig:reasoning_path_effect}
\end{figure}


In this subsection, we fine-tune CodeT5-Base on our reasoning datasets, which are differentiated by the number of reasoning paths they contain, to analyze the effect of reasoning path multiplicity on the reasoning performance of SLMs. This examination aims to discern how the diversity and quantity of reasoning paths in training data influence the model's ability to perform reasoning tasks.


Figure~\ref{fig:reasoning_path_effect} presents the results of our experiments, which demonstrate that a variety of reasoning paths can bolster the reasoning performance of SLMs. For instance, CodeT5-Base, when trained on an ETD dataset featuring four reasoning paths, attains a 38.89\% PoT accuracy on the GSM8K test dataset and a 44.13\% EoT accuracy on ASDiv. In contrast, CodeT5-Base trained on an ETD dataset with only one reasoning path achieves the same 38.89\% PoT accuracy on GSM8K but a lower 31.87\% EoT accuracy on ASDiv. This suggests that the inclusion of multiple reasoning paths in training data can significantly enhance the model's performance, particularly in tasks requiring explanation generation.

\section{Conclusion}
Our research improves how smaller language models (SLMs) think by introducing two new techniques, Equation-of-Thought Distillation (EoTD) and Ensemble Thoughts Distillation (ETD). These methods enable SLMs to perform complex mathematical reasoning. EoTD breaks down reasoning into simpler equations, while ETD uses a range of examples to boost performance. Our findings show that these approaches enhance the SLMs' reasoning skills, making them suitable for use in environments with limited computing resources. However, there are still some limitations hindering the further improvement of our method's mathematical reasoning capability. The first limitation is that LLMs are typically pretrained on natural language and program data. Consequently, when using EoT, they exhibit poorer mathematical reasoning performance compared to CoT and PoT, resulting in a smaller EoTD dataset size. The second limitation is that SLMs are also pretrained on natural language and program data. Thus, fine-tuning SLMs with CoTD and PoTD datasets enhances their reasoning ability, whereas EoTD lacks corresponding SLMs, limiting its potential to improve SLMs' mathematical reasoning performance. We're working on further overcoming these limitations and exploring our method's application beyond mathematics to enhance SLM versatility.

\section*{Acknowledgments}
The work of Jian Li is supported partially by National Natural Science Foundation of China (No. 62106257). The work of Yong Liu is supported partially by National Natural Science Foundation of China (No.62076234), Beijing Outstanding Young Scientist Program (No.BJJWZYJH012019100020098), the Unicom Innovation Ecological Cooperation Plan, and the CCF-Huawei Populus Grove Fund.

\bibliographystyle{elsarticle-num}
\bibliography{ref}{}

\appendix

\begin{figure*}[!ht]
	\centering
	\includegraphics[width=\textwidth]{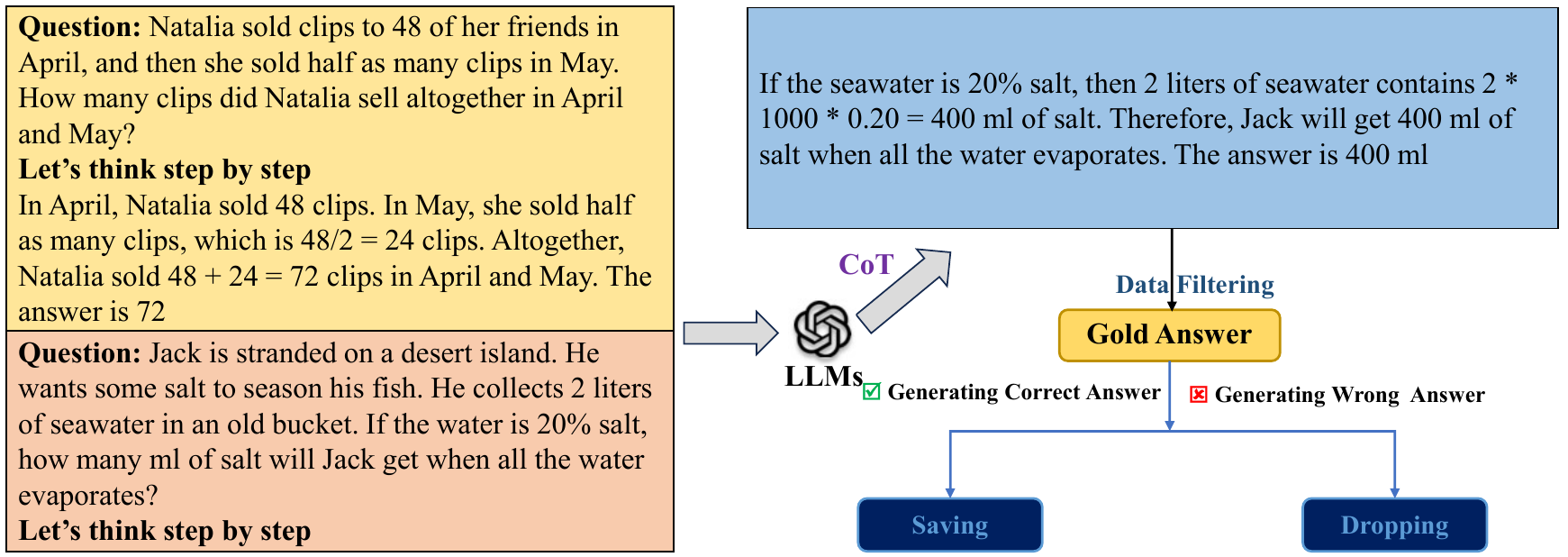}
	\caption{\textbf{Detailed data generation of CoTD.} Firstly, we manually construct some contextualized examples, and combine these contextualized examples, the question, and the prompt ``Let's think step by step" to prompt LLMs to generate CoT based on the question. Then, We extract the answer from this rationale. If the answer doesn't agree with the gold answer, we will drop the CoT. Finally, we get a high-quality reasoning dataset.}
	\label{fig:cot_data_generation}
\end{figure*}

\section{Chain-of-Thought Distillation}
\label{sec:CoTD}
\subsection{Data Generation from LLMs}
\label{subsec:cotdg}
The CoTD process commences with the generation of a dataset from LLMs, which lays the groundwork for subsequent fine-tuning of SLMs. As illustrated in Figure~\ref{fig:cot_data_generation}, we employ in-context learning strategies \citep{abs-2301-00234,min-etal-2022-rethinking,rubin-etal-2022-learning} to elicit the generation of rationales from LLMs within a mathematical reasoning dataset $\mathcal{D}$, where each entry is a tuple $(x,y)$—$x$ being the question and $y$ the correct answer. To generate CoT, we select $k$ samples $\{(x_1, y_1), (x_2, y_2), \ldots, (x_k, y_k)\}$ from $\mathcal{D}$ and manually craft corresponding rationales $c$ in CoT format. These form contextualized examples $\{(x_1, c_1, y_1), (x_2, c_2, y_2), \ldots, (x_k, c_k, y_k)\}$, which are compiled into a demonstration set $\mathcal{D}_D$. We then prompt the LLM with a new question appended with ``Let's think step by step" and feed it the demonstration set to generate a rationale for the question. The CoT generation formula is:
\begin{equation*}
	c_i = f_\mathcal{M}(x_i, \mathcal{D}_D),
\end{equation*}
where $\mathcal{M}$ denotes the LLM, $f$ is the greedy decoding function, and $i$ indexes the example $(x, y)$ in $\mathcal{D}$. This procedure results in a CoT dataset $\mathcal{D}_P$, composed of triplets $(x, c, y)$.

\textbf{Data Filtering}---Upon generating CoT dataset with LLMs, we validate the rationale against the gold answer, a crucial step to ensure the quality of our initial reasoning dataset $\mathcal{D}_C$. Discrepancies between the rationale's answer and the gold answer result in exclusion from $\mathcal{D}_C$. This filtering meticulously purges incorrect instances, enhancing the dataset's quality. Consequently, this refinement directly contributes to the enhanced performance of fine-tuned SLMs, attributable to the increased accuracy and dependability of the training data.

\subsection{Fine-tuning SLMs}
After assembling the reasoning dataset $\mathcal{D}_C$, we fine-tune SLMs on it. For each training instance $(x, c, y)$ from $\mathcal{D}_C$, we prepend the prompt $p_c$ ``Let's think step by step" to the question $x$ to form the input. Fine-tuning is then applied to the SLM to generate the corresponding rationale. The loss function for fine-tuning is:
\begin{equation*}
	\mathcal{L} = - \sum_{i=1}^{N} \sum_{t=1}^{T} \log P(c^i_t \mid c^i_{< t}, x^i, p_c),
\end{equation*}
where $N$ is the number of examples in $\mathcal{D}_C$, $p_c$ is the prompt guiding the SLM to generate the rationale $c$, and $c_{:T}$ represents the sequence of rationale steps $\{c_1, c_2, \ldots, c_T\}$.

After fine-tuning, the SLM becomes proficient at initiating a reasoning process for complex questions, generating the corresponding rationale. The final answer is then extracted from this rationale.

\begin{figure*}[!ht]
	\centering
	\includegraphics[width=\textwidth]{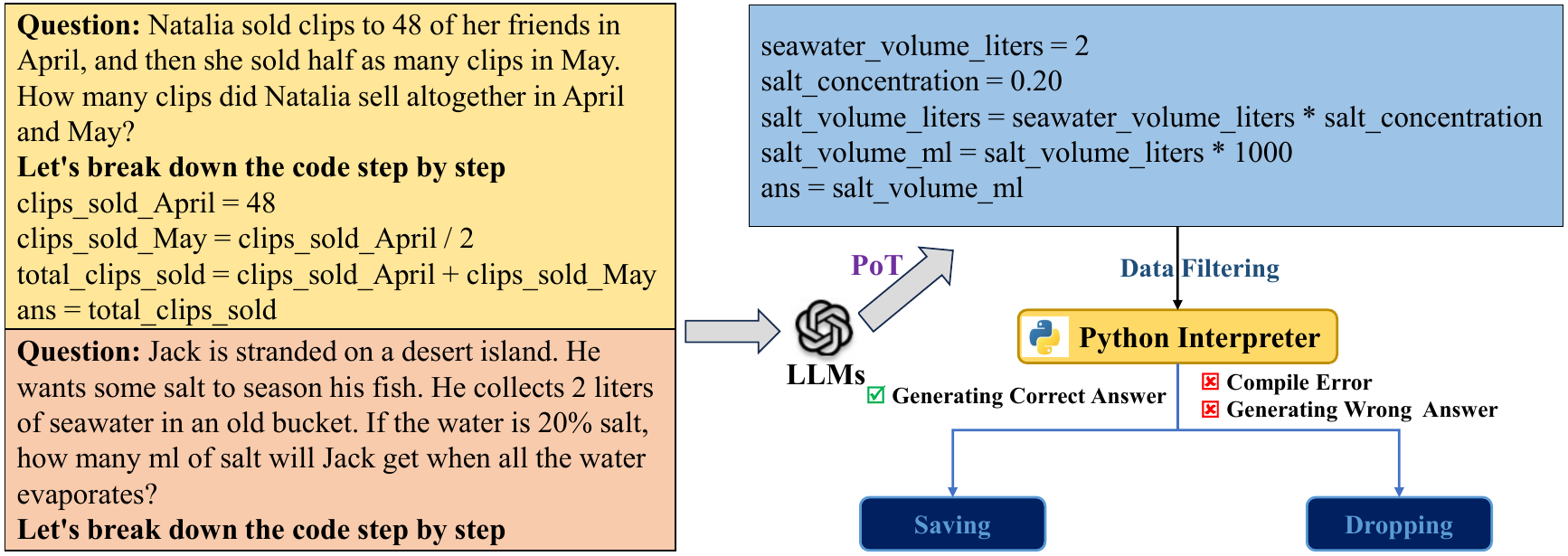}
	\caption{\textbf{Detailed data generation of PoTD.} Firstly, we manually construct some contextualized examples, and combine these contextualized examples, the question, and the prompt ``Let's break down the code step by step" to prompt LLMs to generate PoT based on the question. This program is sent to  a extra python interpreter, if there are compile errors or if it produces wrong answer, we will drop the PoT. Finally, we get a high-quality reasoning dataset.}
	\label{fig:pot_data_generation}
\end{figure*}

\section{Program-of-Thought Distillation}
\label{sec:PoTD}
\subsection{Data Generation from LLMs}
\label{subsec:potdg}
The initial phase in our PoTD entails creating a dataset from LLMs, setting the stage for SLM fine-tuning. Figure~\ref{fig:pot_data_generation} outlines this data generation process. We utilize in-context learning methods \citep{abs-2301-00234,min-etal-2022-rethinking,rubin-etal-2022-learning} to induce LLMs to produce reasoning data. Within the mathematical reasoning dataset $\mathcal{D}$, each entry is a tuple $(x,y)$, where $x$ is the question and $y$ the gold-standard answer. For PoT generation, we choose $k$ samples $\{(x_1, y_1), (x_2, y_2), \ldots, (x_k, y_k)\}$ from $\mathcal{D}$ and manually create rationales $p$ in PoT format. These form contextualized instances $\{(x_1, p_1, y_1), (x_2, p_2, y_2), \ldots, (x_k, p_k, y_k)\}$, which are compiled into a demonstration set $\mathcal{D}_D$. We then prompt the LLM with a new question accompanied by ``Let's break down the problem step by step" and input the demonstration set to generate a rationale for the question. The PoT generation is formalized as:
\begin{equation*}
	p_i = f_\mathcal{M}(x_i, \mathcal{D}_D),
\end{equation*}
where $\mathcal{M}$ is the LLM, $f$ the greedy decoding function, and $i$ the index of the instance $(x, y)$ in $\mathcal{D}$. This yields a PoT dataset $\mathcal{D}_P$, organized as triplets $(x, p, y)$.

\textbf{Data Filtering}---Following PoT dataset generation by LLMs, each program undergoes validation using an external Python interpreter, a vital step to ensure the quality of our initial dataset $\mathcal{D}_P$. Programs that fail to compile or produce incorrect results are immediately discarded. This rigorous filtering process removes flawed instances, thus improving the dataset's quality. The removal of these errors significantly enhances the performance of the fine-tuned SLMs due to the increased accuracy and dependability of the training data.

\subsection{Fine-tuning SLMs}
After generating the reasoning dataset $\mathcal{D}_P$, we fine-tune SLMs on it. For each instance $(x, p, y)$ from $\mathcal{D}_P$, we append the prompt $p_p$ ``Let's break down the code step by step" to the question $x$ for input, and apply fine-tuning to the SLM to produce the corresponding program. The fine-tuning loss function is:
\begin{equation*}
	\mathcal{L} = - \sum_{i=1}^{N} \sum_{t=1}^{T} \log P(p^i_t \mid p^i_{< t}, x^i, p_p),
\end{equation*}
where $N$ is the count of examples in $\mathcal{D}_P$, $p_p$ is the prompt guiding the SLM to generate program $p$, and $p_{:T}$ represents the sequence of program steps $\{p_1, p_2, \ldots, p_T\}$.

After fine-tuning, the SLM excels at initiating a reasoning process for complex questions, producing actionable programs. These are then executed by an external Python interpreter to obtain the final answer.

\begin{figure*}[!hb]
	\centering
	\includegraphics[width=\textwidth]{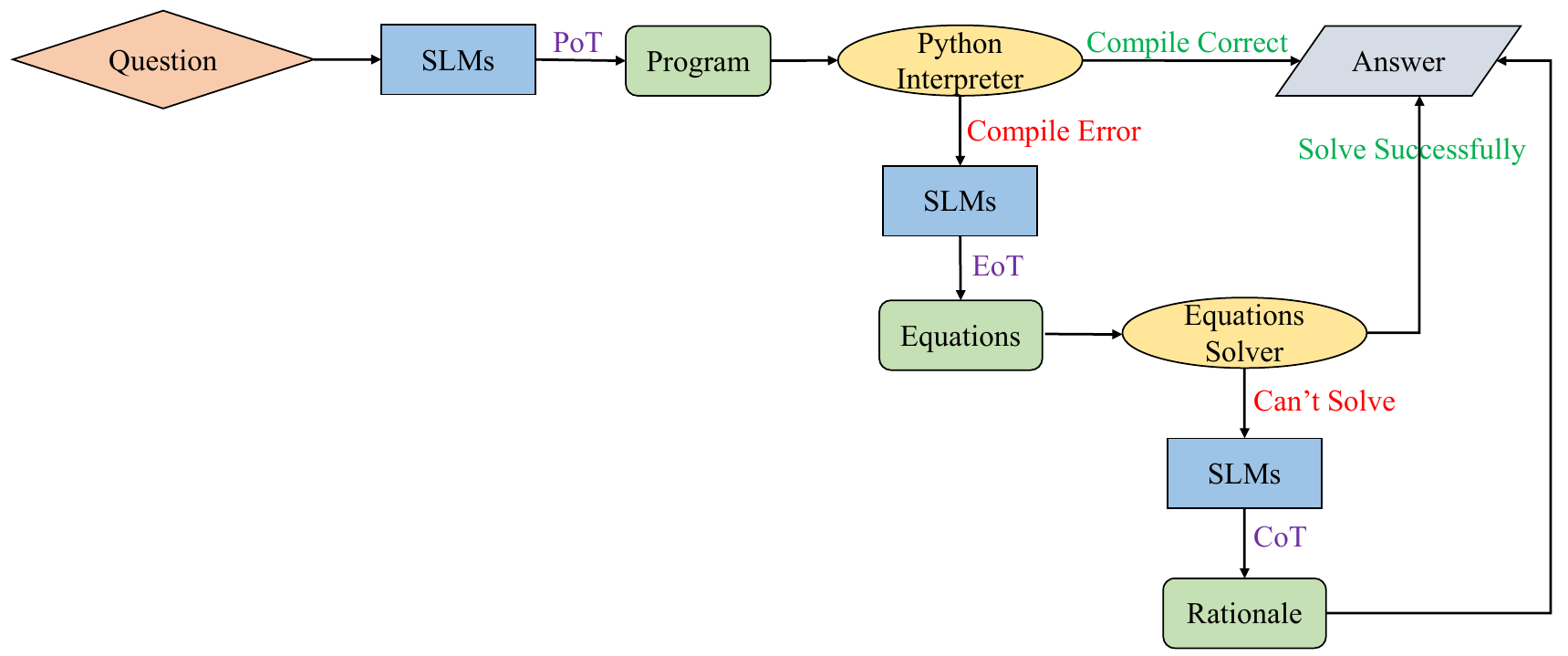}
	\caption{\textbf{The detail reasoning process of ETD.} When given a question, we first prompt SLMs to generate program in PoT, and send the program to an python interpreter to get the final answer. If the program fails to compile, we will prompt SLMs to generate equations in EoT, and send these equations to a equations solver to solve these equations to get the final answer. If these equations can't be solved, we will prompt SLMs to generate rationale in CoT, and extract the final answer from  rationale.}
	\label{inference_ETD}
\end{figure*}

\end{document}